\documentclass[11pt]{article}
\usepackage{cite}
\usepackage{amsmath,amssymb,amsfonts}
\usepackage{algorithmic}
\usepackage{graphicx}
\usepackage{algorithm,algorithmic}
\usepackage{hyperref}
\usepackage{textcomp}
\usepackage{color}
\usepackage{setspace}
\usepackage{booktabs}
\usepackage[margin=1.2in]{geometry}

\newcommand{\MAT}[1]{\mathbf{#1}}
\newcommand{\superscript}[1]{\ensuremath{^\textrm{#1}}}

\renewcommand{\th}{\superscript{th}}

\newcommand{\mx}{\MAT{X}}
\newcommand{\ms}{\MAT{S}}
\newcommand{\ma}{\MAT{A}}
\newcommand{\my}{Y}

\DeclareMathOperator{\std}{std}

\newcommand{\varname}[1]{\texttt{#1}}

\title{Unsupervised Discovery of Clinical Disease Signatures \\
  Using Probabilistic Independence\\

}

\author{Thomas A. Lasko$^1$
\and John M. Still$^1$
\and Thomas Z. Li$^2$
\and Marco Barbero Mota$^1$
\and William W. Stead$^1$
\and Eric V. Strobl$^1$
\and Bennett A. Landman$^2$
\and Fabien Maldonado$^1$}

\date{%
    $^1$Vanderbilt University Medical Center, Nashville, TN\\%
    $^2$Vanderbilt University, Nashville, TN\\[2ex]%
}
\begin{document}

\maketitle

\begin{abstract}
  Insufficiently precise diagnosis of clinical disease is likely responsible
  for many treatment failures, even for common conditions and treatments. With
  a large enough dataset, it may be possible to use unsupervised machine
  learning to define clinical disease patterns more precisely. We present an
  approach to learning these patterns by using probabilistic independence to
  disentangle the imprint on the medical record of causal latent sources of
  disease. We inferred a broad set of 2000 clinical signatures of latent
  sources from 9195 variables in 269,099 Electronic Health Records. The learned
  signatures produced better discrimination than the original variables in a
  lung cancer prediction task unknown to the inference algorithm, predicting
  3-year malignancy in patients with no history of cancer before a solitary
  lung nodule was discovered. More importantly, the signatures' greater
  explanatory power identified pre-nodule signatures of apparently undiagnosed
  cancer in many of those patients.

\end{abstract}

\section{Introduction}
\label{sec:introduction}
A meaningful fraction of treatment failures for common medical conditions may
be due to insufficiently precise diagnoses
\cite{Anderson2008,Ringman2014,Gutmann2014,Tuomi2014,Oksel2018,Gul2019,Schoettler2020,Wang2021}. Existing
diagnostic and treatment protocols have been developed using expert clinical
judgement, biochemical and molecular research, and large-scale clinical
trials. They have served us well for the past century, but higher-precision
treatment decisions will inevitably require higher-resolution diagnostic
clinical pictures.  We view \emph{phenotype discovery} as the work of learning
those pictures from Electronic Health Records (EHRs) using data-driven
computational methods \cite{Lasko2013}.

Our simplified mental model treats a patient's medical record as the result of
a stochastic process that generates elements to appear in the record. The
process is controlled by a set of probability distributions that vary over time
and between patients. When a patient develops a new medical condition, that
condition causes a set of changes to those distributions. We call that set of
changes the \emph{clinical signature} of the condition. We consider the
condition itself to be an unobserved or latent source for those changes, and
multiple latent sources may be active in a given patient at a given time. The
sources do not act directly on the record, but they shape the probability
distributions that govern the data generating process.

For example, a patient who develops type 2 diabetes will experience an
increased probability per unit time that a \varname{type 2 diabetes} billing
code will appear in her record, and a smaller but nonzero increase in the
probability that an (incorrect) \varname{type 1 diabetes} code will
appear. (For clarity, we use an \varname{alternate font} for names of observed
variables.)  The condition may also cause, among other things, changes to the
distribution of \varname{hemoglobin A1C} values, to the distribution of
\varname{blood glucose} values and the frequency of measuring them, and to the
probability that \varname{metformin} will appear as a medication. These
probabilities evolve as the condition progresses, with late-stage disease
producing greater changes in blood chemistry and in code density for end-organ
failure. In early stage disease, elements may appear due to the mere suspicion
of a condition, representing attempts to diagnose or rule out
disease. Representing EHR elements as probabilistic events generated by a
stochastic process allows for these investigative actions, typical
documentation errors, measurement noise, and visit timing to act as evidence
for or against the presence of a particular disease.

\begin{figure}[bt]
  \centering
  \includegraphics[width=\textwidth]{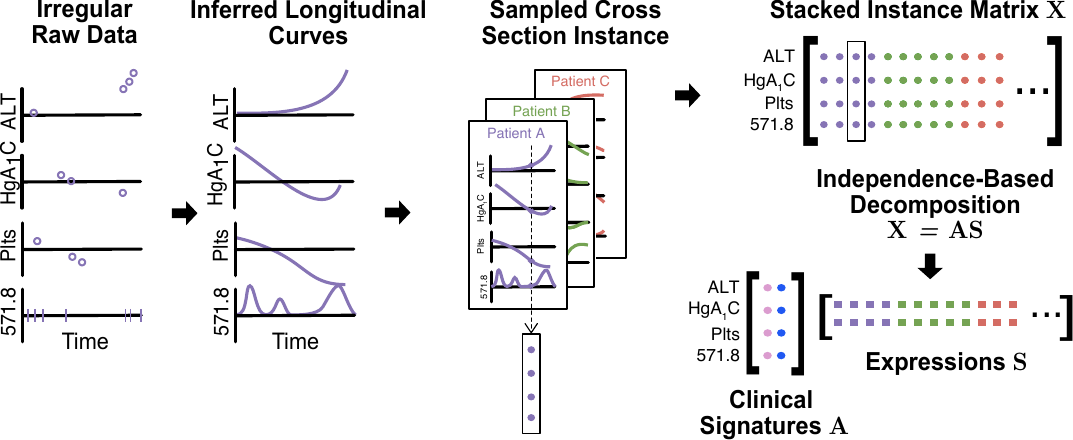}
  \caption{The pipeline for learning clinical signatures and their
    patient-level expressions from noisy, asynchronous, and irregular EHR
    data. The first three steps transform the data into a dense, regular matrix
    $\mx$ for machine learning. The final step infers the clinical signatures $\ma$
    and the expression levels $\ms$ of latent disease sources using probabilistic
    independence.}
  \label{fig:pipeline}
\end{figure}

Representing the relevant probabilities as continuous longitudinal curves
transforms the sparse and irregular EHR data into a form more amenable to
machine learning \cite{Lasko2013,Lasko2019} (Figure \ref{fig:pipeline}). For
example, the discrete events of diagnosis and procedure codes appearing in the
record can be transformed into a continuous intensity curve of a
non-homogeneous Gamma process \cite{Lasko2014}; irregularly measured laboratory
values can be transformed into a continuous Gaussian process that represents a
distribution over all value trajectories that could have produced the observed
data \cite{Lasko2013,Lasko2015}; the discrete decisions to order the test
itself can also be represented as a Gamma process \cite{Lasko2015}; and
medication records can be transformed into continuous curves of total daily
dose, or in cases of incomplete data, curves representing the probability that
the patient took the given medication on a given day. Once the curves are
computed for a given record, a cross section of all of their values at a given
time may be extracted to estimate a patient's clinical state at that moment,
and a matrix of stacked cross sections from all records provides a dense,
regular dataset for further analysis.

The central difficulty in learning clinical signatures comes from the fact that
patients experience multiple simultaneous conditions, so the record is a
confluence of overlapping observations, and the inference task is to
disentangle the active signatures from those observations.

This paper provides an introduction to and real-world evaluation of an approach
to using probabilistic independence to perform the disentangling. It expands on
a preliminary report \cite{Lasko2019}, analyses of the underlying theory
\cite{Strobl2022d,Strobl2023,Strobl2023a}, and some early applications
\cite{BarberoMota2022,Li2023}.

\begin{figure}[tb]
  \centering
  \includegraphics[width=.4\textwidth]{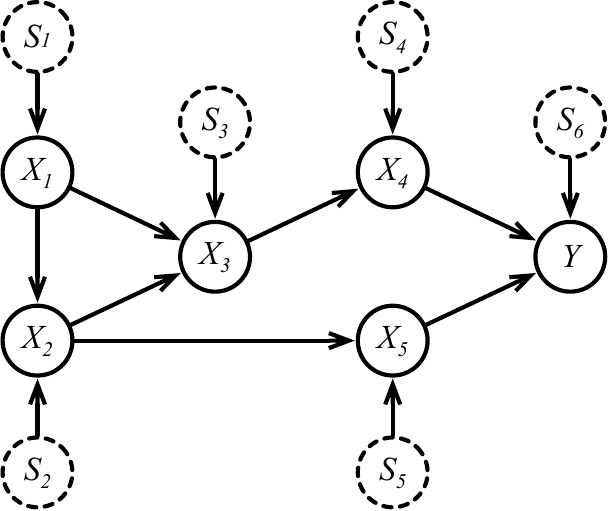}
  \caption{Example causal graph showing observed variables $X_i$, label $\my$,
    and unobserved (latent) sources $S_i$.}
  \label{fig:graph}
\end{figure}

There are both empirical and theoretical reasons why we use this
approach. Empirically, we found early in our work that clinical signatures
identified using independence appear more coherent and clinically recognizable
than those found using other disentangling methods, such as Singular Value
Decomposition \cite{Joliffe2004,Abdi2010,Strang2003SVD}, Latent Dirichlet
Allocation (LDA) \cite{Blei2003}, or various types of deep autoencoders
\cite{Vincent2010,Masci2011a,Doersch2021,Chen2019c,Burgess2018,Mathieu2019}. On
an analogous task in the genomic domain, objectively comparing 42 different
methods of inferring groups of genes that are functionally related and
co-regulated, investigators found that independence-based methods outperformed
all others \cite{Saelens2018}. A different group found that transcriptome
structure inferred using Independent Component Analysis (ICA) was conserved
across five datasets, despite those datasets coming from different research
groups using different technologies \cite{Sastry2021}.

Most importantly, under certain reasonable assumptions of causal inference,
clinical signatures inferred using probabilistic independence have been shown
theoretically to reflect root causes of disease
\cite{Strobl2022d,Strobl2023,Strobl2023a}, at least as far as those causes can
be identified using the variables in the dataset. For example, suppose we have
the causal graph formed by the observed variables $X_i$ and label $\my$ in
Figure \ref{fig:graph}. The value of each observed $X_i$ is a function of its
parents, including an unobserved latent source $S_i$. We can think of the graph
as existing in a steady state until a latent source changes its value, which
then affects downstream variables. Those effects follow a specific pattern
$A_i$ (which we've been calling the clinical signature of the source), that
depends on the presence and strength of causal relationships in the relevant
parts of the network. For example, if source $S_1$ changes, it may affect any
of the $X_i$ and $\my$, but if $S_4$ changes, it can only affect $X_4$ and
$\my$.

The key property of the $S_i$ values (which we call source \emph{expressions})
is that they are all mutually independent (as in Figure \ref{fig:graph}), and
this property is exploited to make the inference. Methods such as ICA
\cite{Hyvarinen2000,Hyvarinen2001} and Linear Non-Gaussian Acyclic Models
(LiNGAM) \cite{Shimizu2006,Shimizu2011} can be used to recover the causal
graph, the source signatures $A_i$, and the source expressions $S_i$ from a
matrix $\mx$ of many instances of observed values. This is done with
unsupervised learning over $\mx$ alone, without the guidance of any label
$\my$.  Under these methods (including our use here), the relationship between
$\mx$ and $\ms$ is assumed linear and additive, so that $\mx = \ma \ms$, but
inference of nonlinear relationships is a topic of current research
\cite{Hyvarinen2019,Hyvarinen2023a,Strobl2023}. The graph and the signatures
remain fixed, with the expressions $\ms$ allowed to vary between patients and
over time, inferred from changes in the observed data $\mx$.

We can build a supervised model to predict a target $\my$ (such as a binary
disease label) from the expressions $\ms$, and use Shapley values
\cite{Lundberg2017,Lundberg2020} to identify the contribution of each source to
the prediction for each patient, accounting for all possible combinations of
predictors. Under this construction, the $S_i$ that produce meaningfully
non-zero contributions should, in theory, represent root causes of $\my$ for
that patient \cite{Strobl2022d}. This is not the case when we use the observed
$\mx$ as the predictive inputs; we can find strong \emph{predictive
  associations} between $\mx$ and $\my$, but in general we cannot make causal
claims using $\mx$ without careful conditioning or other specific analyses.

Phenotype discovery differs from the more common task of \emph{computational
  phenotyping}, in which expert knowledge is used to define computable rule
sets by which known diseases can be recognized in a given patient's record
\cite{Newton2013,Kirby2016,Chapman2021}. The difference is that in phenotype
discovery, we want to \emph{let the data speak}, to tell \emph{us} what the
signatures of all the latent sources are, allowing us to discover emerging
diseases, previously unrecognized conditions, and more precise definitions.

\paragraph{Prior Work}
This research derives from our early work inferring clinical signatures from
EHR data \cite{Lasko2013,Lasko2013b,Lasko2015,Lasko2019}. Much work by others
since then has focused on the broader task of representation learning, in which
features are sought purely for their predictive power, without attempting to
learn clinical patterns reflecting different causes
\cite{Che2015b,Bajor2018,Weng2019,Darabi2020,Steinberg2021,Si2021,Liu2023}.

Some prior work attempts to infer meaningful clinical patterns using other
approaches. Graph Convolutional Transformers have been proposed to learn
functional relationships among diagnosis, treatment, and lab test codes while
simultaneously using the inferred graph for clinical prediction tasks
\cite{Choi2020a}. LDA uses a defined probabilistic graphical model to guide the
decomposition of a data matrix into signature-like `topics'
\cite{Blei2003}. Early work using LDA for phenotype discovery used separately
parameterized distributions for each mode (text, lab orders, billing codes,
medications), which implicitly avoided mode imbalance problems and produced 250
signatures, about 60\% of which were judged by clinicians to be conceptually
coherent \cite{Pivovarov2015}. Later work modified LDA inference to allow for
missing data, and constructed the model in two layers to avoid the imbalance
problem, producing 75 learned signatures \cite{Li2020}. Non-negative matrix
factorization has also been used to decompose matrices of code counts into
small numbers of coarse signatures \cite{Ho2014a,Zhao2019}. Our work differs
from these in the use of probabilistic independence as a guiding principle and
the larger scale of our analysis.

A common approach to phenotype discovery has been \emph{hard clustering}
\cite{Haldar2008,Sutherland2012,Ahmad2014,Shah2015,Li2015c,Zinchuk2020,Hendricks2021a,Landi2020},
in which each patient is assigned to exactly one cluster, using various
measures of patient similarity and cluster quality, with salient clinical
patterns then extracted from each cluster \cite{Jain2010}. Hard clustering is
appropriate when each record legitimately belongs to exactly one cluster, such
as when grouping geographic locations or inferring disease subtypes that are
truly disjoint. However, it fails when trying to identify signals that are
better described as attributes, where each patient expresses varying levels of
multiple attributes simultaneously \cite{Agusti2013}. When misapplied to that
case, hard clustering tends to result in strong attributes overwhelming the
more subtle attributes, which wind up distributed and obscured among the
clusters. In medical records data, strong attribute signals can be
demographics, common conditions unrelated to the disease of interest, or even
which inclusion criteria brought the record into the dataset.

\paragraph{Contributions} In this work we make the following
contributions. First, we demonstrate a scalable approach for transforming
episodic clinical data into continuous, longitudinal probabilistic curves. We
apply the approach to the data modes of clinical measurements, medication
mentions, demographics, and condition billing codes, each of which has unique
behaviors and properties. Second, we demonstrate the use of the principle of
probabilistic independence for large-scale disentangling of clinical
signatures. Third, we evaluate the predictive power of the learned signatures
by using their expression levels to predict the malignancy of solitary
pulmonary nodules, and we assess the stability of our findings under different
random seeds for training. Finally, we investigate the learned signatures'
interpretive power by using them to understand the origin of the pulmonary
nodules in the cohort.

\section{Results}

\subsection{Inference Pipeline}
The inference pipeline included data collection, transformation, and signature
discovery components (Figure \ref{fig:pipeline}). The pipeline is summarized
here, with detailed descriptions in the Methods section.

\paragraph{Data Collection}
We used two datasets drawn from our EHR: a \emph{Discovery Set}, comprising
269,099 records of patients with a broad range of lung disease, used to learn
the latent clinical signatures; and an \emph{Evaluation Set} comprising 13,252
records of patients with no history of cancer preceeding the discovery of a
solitary pulmonary nodule, used to train and test the predictive model. The
Evaluation Set is a subset of the Discovery Set.

We extracted 9195 variables from all records, including clinical measurements,
billing codes, medication mentions, and demographics. Records in the evaluation
set were labeled positive for patients who developed any lung malignancy over
the 3 years after the nodule appearance, and negative for patients who did not
(Table \ref{tab:summary}). 2651 records (20\%) of the Evaluation Set were
randomly partitioned into the final test set.

\begin{table}
  \begin{tabular}{lccc}
\toprule

Attribute            &        Discovery Set & \multicolumn{2}{c}{Labeled Evaluation Set}\\
{}                   &  {}               &        Malignant &       Benign \\
\midrule
  Number of records  &  269,065           &   767           &  12,485           \\
Male sex             &  144,258 (53.6\%) &   371 (48.4\%) &   5,728 (45.9\%) \\
Female sex           &  124,793 (46.4\%) &   396 (51.6\%) &   6,757 (54.1\%) \\
White race           &  202,553 (75.3\%) &   674 (87.9\%) &  10,126 (81.1\%) \\
Black race           &   36,814 (13.7\%) &     63 (8.2\%) &   1,512 (12.1\%) \\
Unknown race         &    17,338 (6.4\%) &     13 (1.7\%) &     449 (3.6\%) \\
Other race           &    12,360 (4.6\%) &     17 (2.2\%) &     398 (3.2\%) \\
Age (years)          &        52 [16,68] &      68 [61,74] &       59 [47,68]
  \\
  \midrule
Per Record\\
\quad Length (days before SPN event) &   1,911 [407,4348] &  1,707 [90,4057] &  2,252 [406,4429] \\
\quad Number condition codes     &       81 [27,224] &     37 [11,125] &      57 [22,149] \\
\quad Number laboratory results           &      210 [41,764] &     87 [14,430] &     139 [39,464] \\
\quad Number medication mentions      &          2 [0,28] &        6 [0,50] &         4 [0,35] \\
\quad Unique condition codes   &        36 [16,71] &      24 [10,51] &       32 [16,61] \\
\quad Unique laboratory tests         &        62 [14,94] &       52 [7,77] &       62 [23,86] \\
\quad Unique medications          &           1 [0,7] &         3 [0,9] &          2 [0,8] \\
\bottomrule

\end{tabular}


  \caption{Data summary statistics. Per record values are Median [IQI]. SPN =
    Solitary Pulmonary Nodule billing code.}
  \label{tab:summary}
\end{table}

\paragraph{Data Transformation}
First, data were converted from episodic point observations to continuous
longitudinal curves with a specific transformation for each data mode, where
the transformation was a scalable approximate version of the more complex but
principled transformation mentioned in the introduction. Clinical measurements
were fit with a smooth interpolation that maintains non-stationarity, or a
constant population median if a patient had no observations of a given
test. Billing codes were represented as a longitudinal intensity curve of code
occurrence events per unit time. Medication mentions were transformed to
piecewise-constant 0/1 curves representing the presence or absence of the
medication in a record at a given date. Demographics were represented as
constant 0/1 curves, dummified from categorical variables when appropriate.

Next, curves for each patient were time-aligned and stacked, with cross
sections sampled uniformly at random at a density of 1 sample per 3
record-years. An individual record may be randomly sampled once, more than
once, or not at all, with longer records tending to be sampled more times. All
sampled cross sections were stacked into a final data matrix of 630,000 rows by
9195 columns. Each column (corresponding to a single variable or
\emph{channel}, such as a specific laboratory test or billing code) was
standardized to place them all onto roughly the same scale.

\paragraph{Clinical Signature Discovery} The matrix $\mx$ of stacked cross
sections was decomposed by FastICA \cite{Hyvarinen2000} into $\mx = \ma \ms$, where
the rows of $\ms$ are mutually independent, column $A_{\bullet i}$ represents the
learned clinical signature of latent source $i$, and matrix element $S_{ij}$
represents the level at which cross section $X_{\bullet j}$ expresses source
$i$. The Direct LiNGAM method \cite{Shimizu2011} is generally a more accurate
method for performing this decomposition \cite{Strobl2022d}, in part because of
the risk of ICA finding a local minimum, but it is far too inefficient for the
matrix size used here.

ICA produces source expressions as rows $S_{i \bullet}$ that are centered at
zero, but arbitrarily ordered and scaled. We re-scaled each row $S_{i \bullet}$
by dividing by two standard deviations, so that roughly 95\% of the values in
each $S_{i \bullet}$ are in $[-1,1]$, and inversely scaled the corresponding
column $A_{\bullet i}$ to maintain the invariant product. But because most of
the $S_{i \bullet}$ are extremely sparse, peaky distributions, and the standard
deviation depends on the degree of sparsity/peakiness, the scaling of the
expressions is not strictly comparable between different signatures.

\paragraph{Clinical Signature Evaluation}
Signatures were evaluated subjectively for clinical coherence, and objectively
by measuring how well they predict the lung malignancy label in the Evaluation
Set.  Elastic Net, Random Forest, and Gradient Boosted Tree (XGBoost)
architectures were trained for the malignancy prediction, one model of each
using signature expressions $\ms$ and another model of each using original
channel values $\mx$ as input. Optimal hyperparameters were determined
independently using cross validation for each of the six models, then training
on the full training set and testing on the held-out test set was repeated with
100 different random seeds each to determine the extent of variation due to
randomness in training. Shapley values \cite{Lundberg2017,Lundberg2020} were
used to interrogate each model for variable importance, and top predictors were
evaluated for the explainability they provide.

\begin{figure}[tbp]
  \centering
  \includegraphics[width=\textwidth]{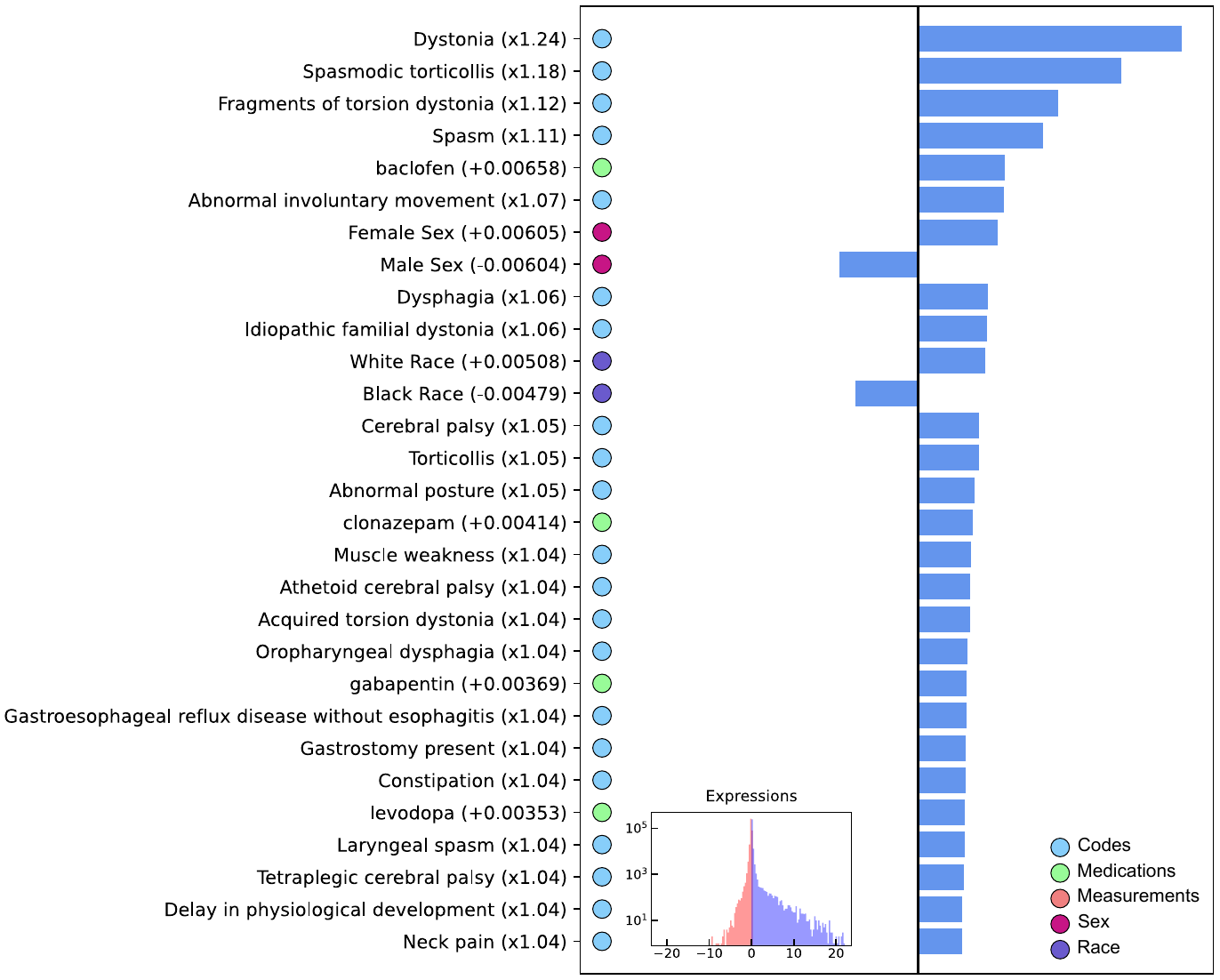}
  \caption{An example clinical signature, selected to illustrate the recovery
    of low-prevalence sources (here, a few hundred out of 630,000 sampled cross
    sections), and not cherry-picked for clinical coherence. We interpret this
    signature as representing the rare condition Spasmodic Torticollis
    \protect\cite{Termsarasab2016}, a specific type of dystonia. Bar length
    gives the size of the change in a given standardized variable, with numbers
    in parentheses indicating the change in original data units for a 1.0
    change in expression. Changes may be either multiplicative
    ($\times \cdots$) or additive ($+ \cdots$). Inset is a log-scaled histogram
    of expression levels in the Discovery Set sample matrix. Expression units
    are individually scaled for each signature such that the standard deviation
    is 0.5, placing 95\% of all expressions within the interval $[-1,1]$.}
  \label{fig:sig-example}
\end{figure}

\subsection{Learned Clinical Signatures}
Signatures of 2000 latent sources were computed and evaluated using the
inference pipeline. Under informal expert review, the learned signatures appear
clinically coherent, representing identifiable clinical pictures, including
rare ones (Figure \ref{fig:sig-example}). The exact proportion of incoherent
signatures was not quantified, and would be subject to disagreement on what
constitutes incoherent, but subjectively the fraction of incoherent signatures
appears to be small. (Signatures for all 2000 latent sources are given in
supplemental material.)  The algorithm assigns only a sequential identifier to
each signature, but for ease of exposition, we have assigned descriptive names
to those relevant to the discussion.

Signatures identify a quantitative pattern of changes to the record made by a
given latent source. These signatures are extremely sparse, with the vast
majority of the 9195 variables indicating standardized changes near zero for a
given source. Our visual representation sorts the changes to all variables and
presents those with an effect greater than a minimum threshold, or the top 10
affected variables, whichever is larger. In Figure \ref{fig:sig-example}, we
interpret the signature as a relatively rare condition of Spasmodic
Torticollis, a specific form of Dystonia.  Quantities in parenthesis give the
effect in original data units of a one-unit increase in expression of the
source. For the \varname{Abnormal involuntary movement} code, this is given as
a factor 1.07, or a 7\% increase in the code intensity. If the expression level
were 10.0 (which from the inset histogram we can see occurs in around 100 of
the samples in the Discovery Set), then this would correspond to code intensity
increase by a factor of $1.07^{10.0} = 1.96$. The signature also specifies a
slightly increased probability that the patient is \varname{female} (increased
by $0.006 \times 10 = 0.06$ for an expression of 10.0) or \varname{white race},
and an increased probability that the patient is treated with
\varname{baclofen}, \varname{clonazepam}, \varname{gabapentin}, or
\varname{levodopa}. All of these align with what is known about the
condition. Botulinum toxin, which is also used to treat the condition, is
omitted here because it fell below the prevalence threshold for our input data,
and so was not included in the analysis.

\begin{figure}[tbp]
  \centering
  \includegraphics[width=\textwidth]{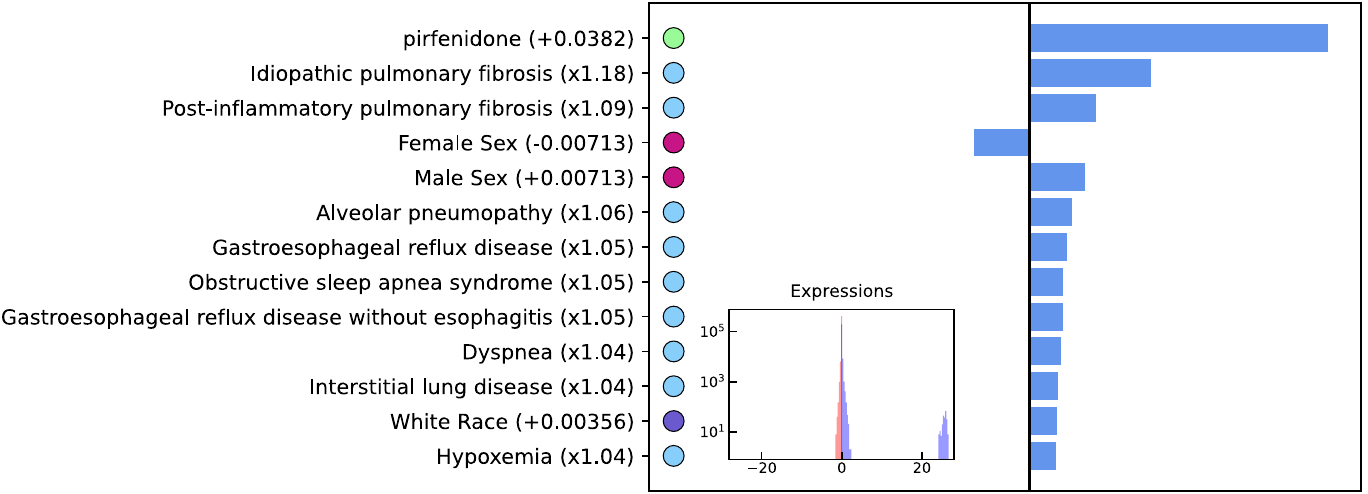}
  \caption{A second example signature of a rarely-expressed source, which we
    interpret as \varname{Idiopathic pulmonary fibrosis} that is treated with
    \varname{pirfenidone}. Several other signatures relate to this same
    condition without the treatment.}
  \label{fig:sig-example2}
\end{figure}

A second example is shown in Figure \ref{fig:sig-example2}. This is for a much
more sparsely expressed source, which we interpret as \varname{Idiopathic
  pulmonary fibrosis} treated with \varname{pirfenidone}. The source is
meaningfully expressed only in some tens of cross sections, all of them with
expression levels around 25. The signature is dominated by the
\varname{pirfenidone} treatment (with a $0.0382 \times 25 = 0.96$ probability
increase of the medication appearing in the record), and includes the known
(but incompletely understood) comorbidities of \varname{Gastroesophageal reflux
  disease} and \varname{Obstructive sleep apnea syndrome}. Not shown are five
other signatures that include \varname{Idiopathic pulmonary fibrosis}, one
(1611, see Supplemental material) that looks like the common clinical picture, one
(467) that explicitly distinguishes \varname{Idiopathic pulmonary fibrosis}
from \varname{Post-inflammatory pulmonary fibrosis}, and three (674, 1113,
1957) that include it as a lower-probability element of a more general picture.

\subsection{Lung Malignancy Prediction}
\begin{figure}[bt!]
  \centering
  \includegraphics[width=\textwidth]{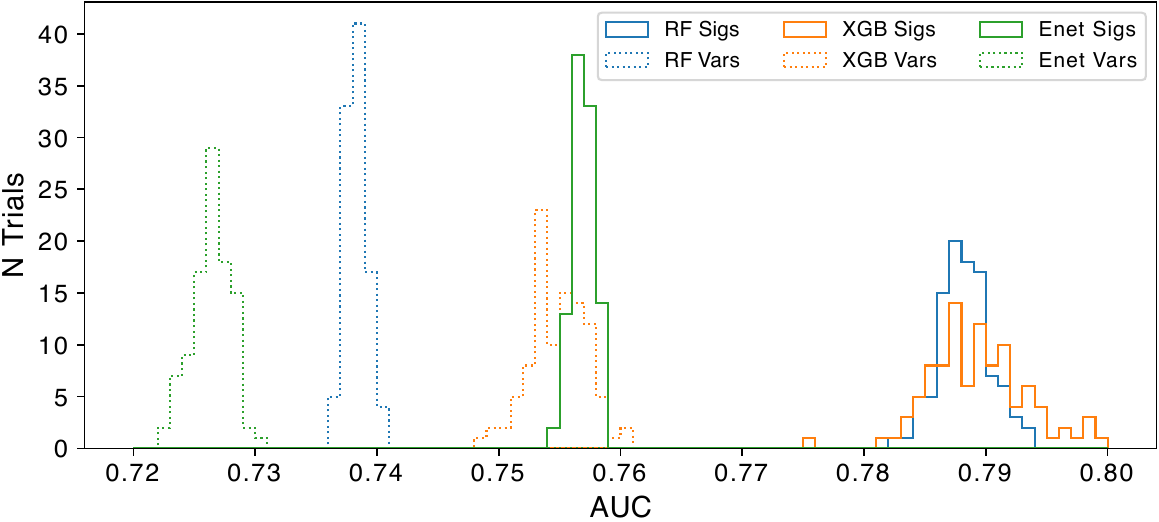}
  \caption{Histograms of area under the ROC curve (AUC) on the held out test
    set for all six models, each retrained 100 times with different random
    seeds. Models using the learned signatures $\ms$ (solid lines) were more
    predictive than those using the observed data matrix $\mx$ (dotted
    lines). Model variability solely due to the choice of random seed was
    substantial, especially for the XGBoost models. Median performance is a more
    robust indicator of performance; extremes in the tails are unlikely to
    generalize to unseen data \protect\cite{DAmour2022}.}
  \label{fig:auc}
\end{figure}

Models that used signature expressions $\ms$ as input performed uniformly better
at malignancy prediction than those that used the original data $\mx$ (Figure
\ref{fig:auc}), suggesting that the signatures captured additional statistical
strength from the larger Discovery Set, despite the fact that there were fewer
of them (2000 signatures vs.\ 9195 variables).

Perhaps more important than the improved performance of the signatures as an
input representation is the improved explanatory power they provide. We use the
Random Forest results to illustrate, but the general conclusion holds for all
models.

\begin{figure}[tbp]
  \centering
  \includegraphics[width=.88\textwidth]{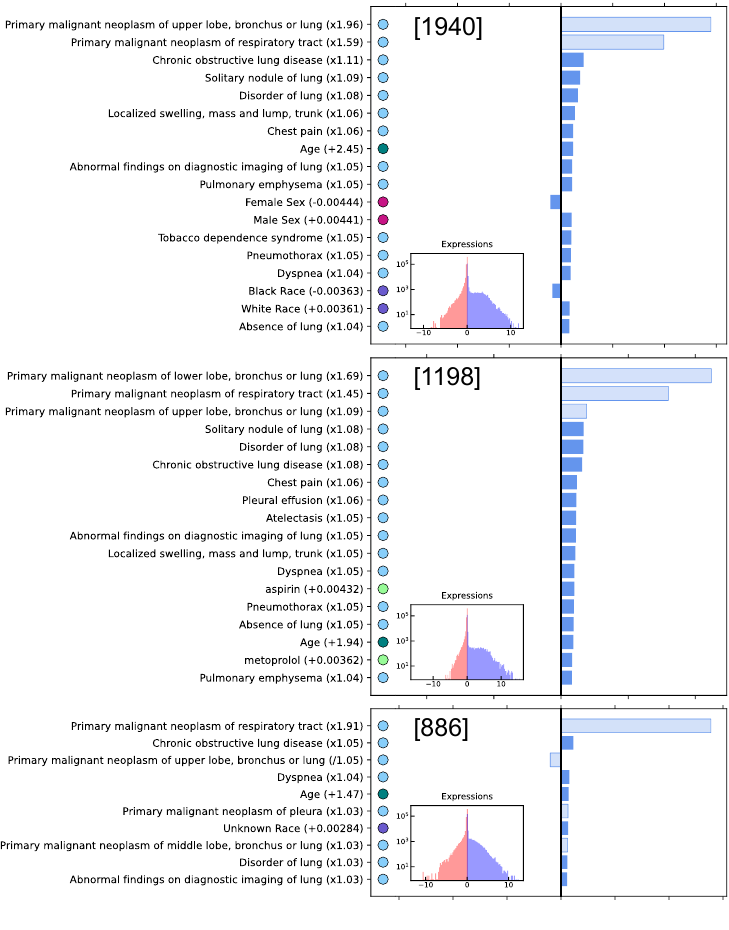}
  \caption{The top 3 predictive signatures in the Random Forest model. We
    interpret these as representing undiagnosed lung cancer for this
    dataset. Light-shaded bars are condition codes for lung malignancy that
    were present in Discovery Set records, but any record in which these codes
    occurred before the first solitary lung nodule code was deliberately
    excluded from the Evaluation Set. The fact that these signatures were top
    predictors for malignancy in that set, despite these missing malignancy
    codes, suggests that the remaining pattern in the signature was
    sufficiently sensitive for malignancy that any patients expressing them had
    undiagnosed cancer.}
  \label{fig:rf-sigs}
\end{figure}

The top three predictive signatures for the Random Forest were signatures of
malignancy in different locations of the respiratory tree (Figure
\ref{fig:rf-sigs}), which makes sense from a causal perspective, but there is
an interesting twist that suggests that they represent undiagnosed disease. The
signatures include as dominant elements billing codes for cancer in various
parts of the respiratory tract. The signatures were learned from the Discovery
Set, which included records of many patients with diagnosed disease that
contained such codes. However, the Evaluation Set was defined to include only
patients with \emph{no history of any type of cancer} before the pulmonary
nodule was discovered, so no record in that set contained any cancer
codes. Nevertheless, the remaining pattern of these signatures, even with the
cancer codes absent in the patient record, turned out to be sufficiently
sensitive that they were important predictors of malignancy, suggesting that at
least some of these records were presenting undiagnosed disease. In practical
application, high-magnitude expression of these signatures for a given patient
record could prompt a search for the relevant malignancy, even before a
solitary nodule appeared on imaging.

The fifth most predictive signature was primary bladder cancer (1452,
Supplemental information), with the same reasons to believe that the model
picked up undiagnosed disease. In this case, it suggests that the lung nodule
in these patients is a metastasis from the bladder.

\begin{figure}[tbp]
  \centering
  \frame{\includegraphics[width=\textwidth]{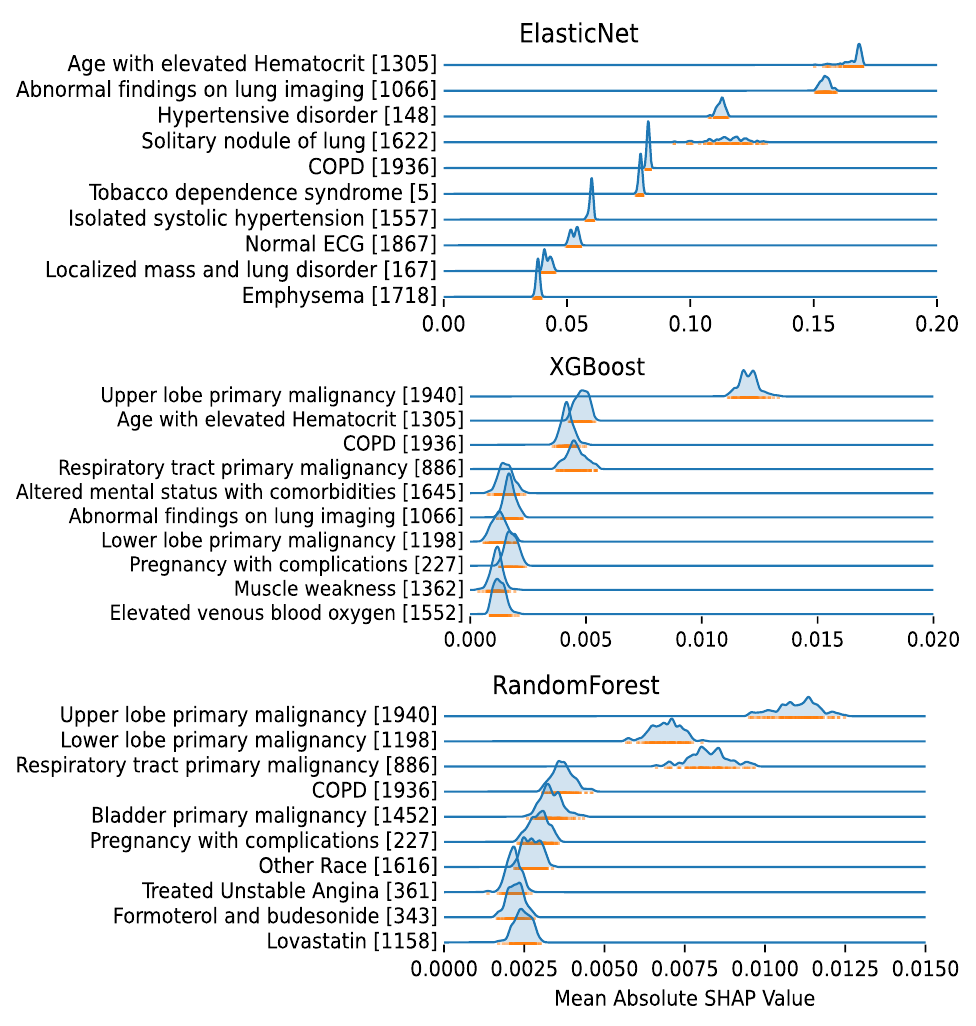}}
  \caption{Predictive importance of the top 10 learned signatures. These are
    distributions over 100 different training random seeds for the optimal
    hyperparameter configuration of each architecture. Row order is by rank
    within the best of the 100. Numbers in brackets are the signature
    identifier in the supplementary material.}
    \label{fig:sig_ridgelines}
\end{figure}

\begin{figure}[tbp]
  \centering
  \frame{\includegraphics[width=\textwidth]{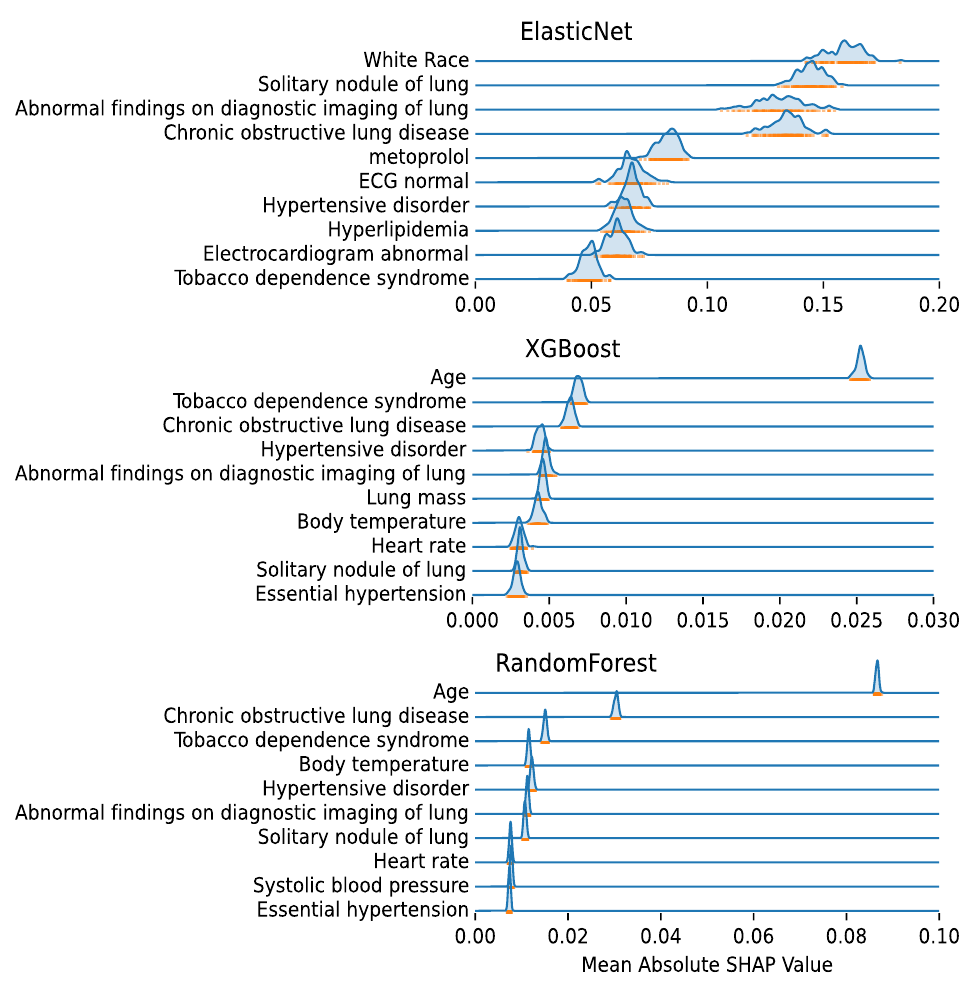}}
  \caption{Predictive importance of the top 10 original variables. These are
    distributions over 100 different training random seeds for the optimal
    hyperparameter configuration of each architecture. Row order is by rank
    within the best of the 100.}
  \label{fig:chan_ridgelines}
\end{figure}

The most predictive latent sources identified by the three architectures were
all informative and plausible (Figure \ref{fig:sig_ridgelines}), although the
models disagreed on specifics of source importance and ranking. In contrast,
the top predictors for models trained on the original data $\mx$ (Figure
\ref{fig:chan_ridgelines}) did not produce actionable insights. Those top
predictors include Age, COPD, Hypertension, and Tobaccco Dependence, all of
which are known strong associations, but none of them is anywhere near as
acutely treatable as a signal of undiagnosed cancer.

The intensity of the \varname{Solitary nodule of lung} code in a record (Figure
\ref{fig:chan_ridgelines}) was a top predictor in the raw-variable models that,
due to how the Evaluation Set was constructed, represented the length of the
record preceeding the first event of that code.  Specifically, each record in
the set has exactly one instance of that code, because that was an inclusion
criterion, and each record was truncated on the date of that code. The curve
builder in our pipeline computed the time-intensity of all codes, which is
usually a time-varying curve, but when any code had a number of events $m < 3$
in a given record of length $l$, our curve builder computed a constant value
$m / l$ for the full curve. Thus, the intensity for this channel became a
constant $1/l$ for each record, and record length is clearly distributed
differently between positive and negative cases (Table \ref{tab:summary}). This
would have represented an information leak had the length of the record been
calculated to include future data beyond the first nodule code, but that wasn't
the case here. Instead, this variable represents a (probably site-specific)
process signal \cite{Lasko2023a} encoding the fact that patients who turn out
to have a malignant nodule tend to have a shorter history at our institution
than patients who turn out to have a benign nodule.

\section{Discussion}

We have described an unsupervised method of using probabilistic independence to
disentangle from EHR data the clinical signatures of multiple, simultaneously
expressed sources of disease. This includes a method to transform the noisy,
sparse, and irregular EHR data of multiple different data modes (billing code
events, medication mentions, clinical measurements, and demographics) into
dense, regular data that can be operated on by standard machine learning
methods.

We demonstrated that a given patient's expression $\ms$ of the latent sources
is more predictive for lung cancer than the dense matrix $\mx$ of original
variables, which implies that the signatures are capturing meaningful clinical
patterns, and may be transferring some statistical strength from the large
unlabeled Discovery Set to the smaller labeled training set.

We investigated the clinical utility of the learned signatures in explaining
the models' predictions and in understanding the pathophysiologic sources of
the predicted lung cancer in patients with a documented solitary lung
nodule. We found that the top predictors display clinical face validity, and
help to understand the causes of malignancy of a given nodule; in this case
there is evidence that the most important causes are previously undiagnosed
cancer, either in the lung or other organs.

Our methods can be extended to include other data sources. In particular,
concepts extracted from progress notes, panomics data, environmental exposures,
and personal sensors could be added to give richer, higher-resolution
signatures without modifying the underlying algorithms.

While in theory we expect the predictors in $\ms$ to reflect root causes of
disease, in practice we found important differences in how the $S_i$ were used
between Elastic Net, Random Forest, and XGBoost models (Figure
\ref{fig:sig_ridgelines}), which urges caution in drawing causal
conclusions. We speculatively attribute the performance difference (about 0.03
AUC) between the linear and tree ensemble models to residual interactions
between the $S_i$ and their potential nonlinear relationship to $Y$, both of
which can be modeled better by tree ensembles.

From our perspective, the Random Forest results look clinically the most
plausible. We hypothesize that the broader and more random search performed by
Random Forest training may have produced a more representative model, while the
greedy XGBoost approach may be more likely to find local minima. Some evidence
for this interpretation lies in the disproportionately dominant use by XGBoost
of \varname{Upper lobe primary malignancy}, the most common location for lung
malignancies. The $10 \times$ higher Shapley values for the top XGBoost
predictors also suggests that the algorithm is anchoring on just a few strong
sources, as we might expect from a greedy algorithm. Understanding these
differences is a direction for future study.

In addition to the variation in the learned structure, the random variation of
discrimination performance was higher than anticipated. The range of AUC under
random-seed variation was about 0.02 for the grandient-boosted tree classifier,
0.01 for the random forest, and 0.005 for the elastic net (Figure
\ref{fig:auc}). This variation was larger than we expected, typically at least
as large as the change due to a hyperparameter adjustment during tuning, which
could easily mislead an algorithmic search for the best hyperparameters. In our
experience, sequential model-based optimization strategies such as those using
Bayesian Optimization \cite{Bergstra2011,Bergstra2013} can be fooled by
randomness this large into settling on non-optimal settings, even though the
optimization algorithms are designed to handle some amount of noise. In
contrast, random search is non-sequential, allowing one to identify an optimum
by eye, after the trials are run. So while sequential optimization can produce
superior models under a given tuning budget, in practice the benefit may depend
on the degree of variation in performance due to the random seed.

The variation under random seed selection also urges caution with the
increasingly common practice of trying many random seeds and then accepting the
best performing model to report. If the goal of a project is to find the single
best performing model for a given purpose (such as for a competition or a
purpose-built model for deployment), then random-seed hunting can be a useful
practice. But if the goal is to evaluate a \emph{given architecture} or a
\emph{given set of predictive variables}, or to understand whether a
\emph{given learning approach} produces better results than some alternative (a
common goal in the machine learning literature, and our goal here), then seed
hunting can produce invalid conclusions, because the effect of the random seed
can overshadow or even reverse the actual performance increment produced by the
idea being tested.

Moreover, seed hunting may backfire when there is any out-of-distribution (OOD)
component of the application, such as when the model is deployed at a different
time or at a different institution from where it was trained
\cite{DAmour2022}. Our results demonstrate that different random seeds can
produce meaningfully different model structure. This variation may either help
or hurt when applied to slightly OOD data, and it is difficult to predict in
which direction that will go \cite{DAmour2022}.

\paragraph{Limitations} In this work we do not demonstrate that any of the
predictive algorithms recover the truth about the source of a given patient's
disease, and the fact that the three models disagree on the global results
demonstrates that at least two of them are incorrect in many
cases. Establishing truth, and investigating the effect of model architecture
on the recovery of that truth, would require a large-scale synthetic data
experiment, and is a direction for future work. Previous experiments in the
domain of liver failure, however, produced results that closely align with what
is known about the disease \cite{Lasko2019}.

We use binary variables for medications and demographics, and interpret the
quantities that represent them in signatures as probabilities. This seems to
work reasonably well, despite the fact that the ICA decomposition
$\mx = \ma \ms$ is defined only for continuous-valued variables. There is
evidence suggesting that binary observations may be used without algorithm
modification if the observations are sparse (mostly zero) or the latent sources
are sparse (mostly near zero) \cite{Himberg2001}. In our case, the medication
variables are sparse, as are nearly all of the latent sources.

Finally, latent signatures $\ma$ are not likely to be transportable, because
the probabilities represented in them capture site-specific practices that are
not likely to be reproduced at a different institution, or even by the same
institution over time \cite{Lasko2023a}. However, to the extent that the
signatures capture latent sources that are intrinsic to disease states, rather
than operational practice patterns, the \emph{expressions} $\ms$ of the latent
sources may play a role in transportability, despite their signatures differing
across sites.

\paragraph{Conclusion} We have demonstrated large-scale, data-driven inference
of latent disease sources, using routinely collected, multi-modal EHR data. The
method recovers coherent, clinically recognizable signatures or patterns of
pathophysiologic changes caused by those sources, including some found in less
than one in $10^{4}$ training instances. The signatures cover a broad range of
disease, and provide intuitive dimensions for understanding a patient
record. In a lung cancer prediction task, they outperformed the original
observed variables as inputs, and suggested clinically undiagnosed malignancy
as top predictors.

\section{Methods}
This research was approved as non-human-subjects research by the Vanderbilt
University Institutional Review Board (\#210761).

\subsection{Data}
All data were extracted from the research mirror of VUMC's EHR, which covers
about 3 million patients, with nearly all inpatient and outpatient records
complete after 2005. HIPAA-defined Personal Health Information (names, record
numbers, absolute dates, etc.) was removed before processing.

Original ICD-9 and ICD-10 billing codes were mapped onto the SNOMED condition
code taxonomy, original medications were resolved into ingredients, and extreme
values for clinical measurements that were deemed impossible or incompatible
with life were removed. Any resulting variables with fewer than 1000 total
observed instances in the database were excluded, and measurements present in
fewer than 10 records of the Discovery Set (see below) were excluded. A final
total of $p=9195$ variables were recorded, including 7453 SNOMED condition
codes, 771 medication ingredients, 949 clinical measurements, and 12
demographics (1 age continuous variable, 3 sex categories, and 8 race
categories).  Categorical variables were dummified into binary variables.

\subsubsection{Discovery Set} A set of 95 condition concepts was curated by a
pulmonologist to include a broad scope of infectious, malignant, and other lung
conditions. Patient records were included in the Discovery Set if they
contained at least one billing code descended from any of the concepts prior to
the cutoff date of 4/29/2022.  Records were excluded if they contained fewer
than two distinct dates containing an observation of a condition, measurement,
or medication. A total of 269,099 records met these criteria.

\subsubsection{Evaluation Set} A cohort of 13,252 records was collected
comprising those with at least one billing code for a solitary pulmonary nodule
(SPN), and no codes for any type of malignancy preceeding the SPN date. Because
the code for an SPN was among the lung condition codes defining the Discovery
Set, the Evaluation Set was a subset of the Discovery Set.

Evaluation Set records were labeled positive if they contained at least one
code for a malignant lung neoplasm (including primary lung cancer or metastasis
from other locations) on day 4 - 1095 following the SPN date. Records were
labeled negative if they had no lung malignancy code before day 1095, even if
they had a malignancy in some other organ after the SPN date. Records with lung
malignancy codes within 4 days of the SPN code were excluded because they
represented cases where the malignancy status was presumed to be known at the
time of the SPN detection. Labels were validated by comparison with cancer
registry records (all positive labeled records) and manual chart review (all
mismatches with the cancer registry and a random sample of negative labeled
records) as detailed elsewhere \cite{Li2023c}, and estimated to have 0.98 PPV,
0.99 NPV, 0.93 sensitivity, and 0.996 specificity.

A random partition of 2651 records (20\%) of the Evaluation Set were set aside
as the final test set.

\subsection{Continuous Curve Generation}
Continuous longitudinal curves were built with methods specific to each data
mode. A curve was generated for each \emph{channel} (corresponding to a single
clinical variable, such as a specific laboratory test or billing code) with
one-day resolution. The time constant of the actual information captured by the
curve was usually not as small as one day, because observations were made in
general at far larger intervals. Nevertheless, creating the longitudinal curves
at this resolution allows us to estimate the value of a given variable on
arbitrary dates, given the set of observations.

\subsubsection{Clinical Measurements} Curves for clinical measurements (mostly
laboratory test results) were originally built using non-stationary Gaussian
process regression \cite{Lasko2015}, which provides a distribution over all
possible paths that the measurand could have taken over time, given the
observed values and assumptions on the smoothness of the path. However, that
method is too inefficient for large-scale use. For this work, we used the
univariate PCHIP method\cite{Fritsch1984} that interpolates a curve with a
continuous first derivative through all observations. This method adapts to
non-stationarity, maintains monotonicity given monotonic data, and does not
overshoot the maxima defined by the observations. Extrapolation beyond the
first and last observation used the value of the nearest observation.

If a record listed no observations for a given clinical measurement, a constant
curve at the population median was imputed.

\subsubsection{Billing Codes} Curves for billing code events were originally
computed as the intensity curves of non-homogeneous gamma processes
\cite{Lasko2014}, which provide a distribution over all possible intensity
curves, given smoothness assumptions, as well as a distribution over a shape
parameter that models dependence between adjacent events.

To trade approximation for complexity with this data mode, we replaced the
gamma process with a simple density function that assumes memorylessness
between events, but is still able to model nonstationarity over
time. Specifically, we used a variation we developed for Random Average Shifted
Histograms (RASH) \cite{Bourel2014}. RASH estimates an intensity curve by
averaging many equally-spaced histograms over the data, with each histogram
shifted by a random amount. It produces a smooth curve similar to kernel
density estimation (KDE), but the constant bin width causes the same problems
with nonstationarity that a constant KDE bandwidth causes. To accommodate
nonstationarity, we modified RASH to use random-sized bins, with the random
choice of bin size made in event space rather than data space. For example, a
bin size of 5.7 would mean that the bin includes the next 5 events plus 70\% of
the interval between the 5th and 6th events. Bins were specified in increasing
order by event time, with each bin size chosen uniformly at random between 3.0
and the number of remaining events. This naturally handles the non-stationarity
in event density, with events spaced further apart in time allocated into wider
time bins than events closer in time.

\subsubsection{Medication Mentions}
Curves for medication mentions were binary curves intended to reflect whether
the patient had been prescribed to take the medication at the indicated
time. Creating the curve was challenging, however, because for much of the
history of our EHR, the start and stop dates of medications are either missing
or unreliable, and dispensing or refill data was not available. Instead, at
some patient visits, a clinician would review the current medication list with
the patient, confirming or removing entries. The modified list was then
inserted into the database and marked with the date of the reconciliation.

Curves were computed by setting them at 1 (taking) for all medications
mentioned on a given reconciliation, and 0 (not taking) otherwise. Times
between reconciliations used the value of the nearest observation.

If a record included no mentions of a given medication, a constant curve of 0
was imputed.

\subsubsection{Demographics} Curves for demographics were either constant
(race, sex) or linearly increasing (age). Categorical values were
dummified into sets of binary curves.

\subsubsection{Smoothing} We found that capturing a small amount of history in
the curves was useful for capturing short-term trajectories of disease and
filling short documentation gaps. To do this, a one-year retrospective rolling
mean was applied to each curve. For medications, the `taking' region of each
curve was extended for 365 days in each direction.

\subsection{Cross-section Sampling} All curves from the $i\th$ record were
aligned into a curveset, anchoring on the first date of the record. A number
$c_i$ of cross-sections were sampled uniformly at random from each record in
the Discovery Set, where $c_i \sim \operatorname{Bin}(l_i, d)$, $l_i$ is the
record length in days and the sampling density $d = 1/(3 * 365)$ was fixed at
one sample per three record-years.  Longer records therefore had a higher
probability of being sampled multiple times, and some records were not sampled
at all. A total of $n=630,037$ cross sections from 175,711 records were sampled
and stacked into the matrix $\mx \in \mathbb{R}^{p \times n}$. The size of
$\mx$ (controlled by sampling density $d$) was limited by available RAM and the
need for the FastICA implementation to hold multiple copies of intermediate
matrices in memory at once.

In the Evaluation Set, records were sampled exactly once, at the time of the
first SPN billing code. No data observed after the SPN date were used to
construct the curves for the Evaluation Set.

\subsubsection{Standardization} The discovery matrix $\mx$ was transformed to
bring variables from all modes onto roughly the same scale. Different
transformations were used for different row subsets that correspond to data
modes.

The row subset $\mx^C = \mx_{i \bullet}, i \in V^C$ of clinical measurement
variables $V^C$ and the row subset $\mx^M = \mx_{i \bullet}, i \in V^M$ of
medication variables $V^M$ were each transformed by subtracting the subset mean
and dividing by two standard deviations.

The row subset $\mx^B = \mx_{i \bullet}, i \in V^B$ of billing code intensity
variables $V^B$ was first transformed to
$\tilde{\mx}^B = \log(\mx^B + \epsilon)$, where the smoothing value
$\epsilon = 1/(20 \times 365))$ represents an arbitrarily chosen prior of one
code per twenty record-years, and then each row $\tilde{\mx}^B_{i \bullet}$ was
scaled by dividing by $s = 2 \cdot \std(\tilde{\mx}^B_{i \bullet})$.

Demographics were not transformed, remaining as 0/1 variables.

The logarithmic transform affects the composition of the signatures and their
expression values, but the scaling operations only affect the relative ranking
of the different data modes in the signature visualizations.

Evaluation set cross sections were standardized using the transformation
determined by the Discovery Set.

\subsection{Clinical Signature Discovery} The final discovery matrix
$\mx \in \mathbb{R}^{p \times n}$ was decomposed by ICA \cite{Hyvarinen2000,
  Hyvarinen2001} into a mixing matrix $\ma \in \mathbb{R}^{p \times k}$ and a
source matrix $\ms \in \mathbb{R}^{k \times n}$, such that $\mx = \ma \ms$, the
rows of $\ms$ are (close to) mutually probabilistically independent, and $k$ is
the number of latent sources inferred. In terms of our application, the column
$X_{\bullet j}$ is the $j\th$ sampled cross section, column $A_{\bullet i}$ is
the clinical signature of latent source $i$, and matrix element $S_{ij}$
represents the level at which cross section $X_{\bullet j}$ expresses source
$i$.

Mathematically, we can infer $k = p$ components, although optimally, we want to
match $k$ to the intrinsic dimension of $\mx$ to avoid over- or under-learning
\cite{Hyvarinen2000}. In our case, we used $k = 2000$. We suspect that the
optimal $k > 2000$, but we were limited by the computational complexity of the
algorithm and the need for the scikit-learn implementation to maintain multiple
intermediate matrices in RAM. To infer $k < p$, standard implementations use
the modified ICA equation $\MAT{V}\mx = \ma \ms$, or
$\mx = \MAT{V}^{-1}\ma \ms = \tilde{\ma} \ms$, where
$\MAT{V} \in \mathbb{R}^{k \times p}$ is obtained from singular value
decomposition and provides the additional benefit of whitening the data
\cite{Hyvarinen2000}. In this paper, when we refer to our computed results as
$\ma$, we are technically referring to this $\tilde{\ma}$.

The matrix $\mx^E$ of the Evaluation Set was created analagously to the
discovery matrix $\mx$, with the exception of each record being sampled exactly
once. The same standardizing transformations were applied (with the same
parameters) as with the Discovery Set. Evaluation Set source expressions
$\ms^E = \tilde{\MAT{A}}^{-1}\mx^E$ were computed, and the matrices $\ms^E$ and $\mx^E$
were partitioned into training and test sets, using the patient-level
partitioning described above.

\subsection{Clinical Signature Evaluation}
First, the face validity of the clinical signatures were evaluated subjectively
and informally, by examining the identified patterns and considering how well
they represent recognizable patterns of disease. Next, they were evaluated
objectively by measuring how well the expression matrix $\ms^E$ predicts the
malignancy label $\my$ in a Random Forest (Python 3.10, scikit-learn 1.1.3), a
Gradient-Boosted Machine (Python 3.10, XGBoost 1.6), and an Elastic Net (Python
3.10, scikit-learn 1.0.2), compared to how well the observed variable matrix
$\mx^E$ predicts $\my$ with the same architectures. Hyperparameters for all six
models were optimized independently using $10 \times$ cross validation using a
combination of random search, grid search, and human guided search. After
optimal hyperparemeters were determined for each model, training on the full
training set and testing on the test set were repeated with 100 different
random seeds each to determine the extent of variation due to randomness in
training.

Models were interrogated for variable importance using Shapley values
\cite{Lundberg2017,Lundberg2020}, which estimate the contribution of each
predictive variable, taking into account all possible interactions of that
variable with others.

\section{Acknowledgements}
This work was funded in part by grant award CA253923 from the National Cancer
Institute.  It relied on data from the VUMC Research Derivative, supported by
CTSA award UL1 TR002243 from the National Center for Advancing Translational
Sciences. Its contents are solely the responsibility of the authors and do not
necessarily represent official views of the National Center for Advancing
Translational Sciences or the National Institutes of Health.

\section{Competing Interests}
The authors declare that there are no competing interests.

\section{Author Contribution}
Overall concept and direction by TAL. Unsupervised learning approach designed
by TAL, WWS, EVS, JMS, and impleneted by TAL, JMS. Supervised evaluation
designed by TAL, BAL, FM, and implemented by JMS, TL, MBM. Analysis and
interpretation of results by all authors. Manuscript draft by TAL, with
critical revision by all authors. All authors take responsibility for the final
manuscript contents.

\section{Data Availability}
The electronic health record data that support the findings of this study are
available upon request from the corresponding author, approval from the
institutions’ respective IRBs, and approval from an additional VUMC
representative assessing institutional risk. Requests for access will be
processed within around 2 months, subject to signing a data use agreement.

\section{Code Availability}
Source code for this project is available at
\url{https://github.com/ComputationalMedicineLab/ipn_methods_code_public}.

\bibliography{None.bib}
\bibliographystyle{vancouver.bst}

\end{document}